\documentclass[runningheads]{llncs}

\usepackage{eccv}

\usepackage{eccvabbrv}

\usepackage{graphicx}
\usepackage{booktabs}

\usepackage[accsupp]{axessibility}  % Improves PDF readability for those with disabilities.

\usepackage[pagebackref,breaklinks,colorlinks,citecolor=eccvblue]{hyperref}
\usepackage{orcidlink}

\usepackage{microtype}
\usepackage{graphicx}
\usepackage[ruled,linesnumbered]{algorithm2e}
\usepackage{multirow}
\usepackage{array}
\usepackage{algpseudocode}
\usepackage{amsmath}
\usepackage{amssymb}
\usepackage{mathtools}
\begin{document}

\title{Heterogeneous Federated Learning with Splited Language Model}

\titlerunning{Heterogeneous Federated Learning with Splited Language Model}

\author{Yifan Shi\inst{1,3,}\thanks{ Equal Contribution. } \and
Yuhui Zhang\inst{2,3,}\thanks{ Equal Contribution. } \and
Ziyue Huang\inst{3}\and 
Xiaofeng Yang\inst{3} \and Li Shen\inst{4} \and Wei Chen\inst{2} \and Xueqian Wang\inst{1}
}

\authorrunning{Yifan Shi et al.}

\institute{Tsinghua University, Shenzhen, China 
% \url{http://www.springer.com/gp/computer-science/lncs} 
\\ 
\and
State Key Lab of CAD\&CG, Zhejiang University, Zhejiang, China
\and
Tencent, Shenzhen, China
\and
JD Explore Academy, Beijing, China
}

\maketitle

\begin{abstract}
Federated Split Learning (FSL) is a promising distributed learning paradigm in practice, which gathers the strengths of both Federated Learning (FL) and Split Learning (SL) paradigms, to ensure model privacy while diminishing the resource overhead of each client, especially on large transformer models in a resource-constrained environment, e.g., Internet of Things (IoT). However, almost all works merely investigate the performance with simple neural network models in FSL. Despite the minor efforts focusing on incorporating Vision Transformers (ViT) as model architectures, they train ViT from scratch, thereby leading to enormous training overhead in each device with limited resources. 
Therefore, in this paper, we harness Pre-trained Image Transformers (PITs) as the initial model, coined FedV, to accelerate the training process and improve model robustness. Furthermore, we propose FedVZ to hinder the gradient inversion attack, especially having the capability compatible with black-box scenarios, where the gradient information is unavailable. Concretely,  FedVZ approximates the server gradient by utilizing a zeroth-order (ZO) optimization, which replaces the backward propagation with just one forward process. Empirically, we are the first to provide a systematic evaluation of FSL methods with PITs in real-world datasets, different partial device participations, and heterogeneous data splits. Our experiments verify the effectiveness of our algorithms.
\end{abstract}

%%%%%%%%% BODY TEXT
\section{Introduction}

Federated learning (FL) \cite{mcmahan2017communication,Li2020fl} allows distributed clients to collaboratively train a shared model under the orchestration of the cloud without transmitting local data in parallel. Meanwhile, Split Learning (SL) \cite{gupta2018distributed,vepakomma2018split} divides a whole neural network into multiple subnetworks to protect the privacy between the server and clients with limited computing resources (e.g., Internet of Things, IoT). Nowadays, these two distributed learning paradigms gradually become promising approaches in real-world applications, such as healthcare \cite{park2021federated, park2022multi,kim2022task,almalik2023fesvibs}, edge devices (e.g., mobile phones) in IoT \cite{hard2018federated,wei2021user,shi2023efficient}, and wireless network \cite{niknam2020federated,li2021blockchain,deng2022blockchain}, where it is vital to protect data privacy.

To consider the limited resources of edge devices and training overhead in terms of training time, computation, and communication costs when introducing large language models (LLMs) into FL methods \cite{lin2021fednlp,guo2023promptfl,yu2023federated,liu2023federated,guo2023pfedprompt,lu2023fedclip}, the emerge of Federated Split Learning (FSL) as a new paradigm extensively appeals to many researchers \cite{singh2019detailed,gao2020end,huang2023minibatch,han2021accelerating,mu2023communication,park2021federated,almalik2023fesvibs}.
It provides model privacy and reduces the training overhead, especially for each client in resource-constrained environments by splitting the whole model into two portions: multiple client-side models residing in local clients and a server-side model stored in the server, meanwhile, adopting the parallel training paradigm between the client and server similar as FL instead of sequential training in SL.

However, similar to FL, FSL also suffers from the data heterogeneity challenge in practical data distributions (e.g., non-IID), which can induce convergence and model weight divergence \cite{li2021model,li2020federated} as well as severe catastrophic forgetting issues \cite{gupta2021addressing, chang2018distributed} for FL methods. Recent works \cite{qu2022rethinking,zuo2022empirical} give insight that simply replacing convolutional neural networks (CNN, e.g., ResNet model family) with Vision Transformers (ViTs) can greatly improve the model performance in FL, especially more robustness against heterogeneous data distributions due to being capable of encoding long-range dependencies between input sequences \cite{qu2022rethinking,khan2022transformers}. Furthermore, the model architecture of ViTs is a straightforward configuration to decompose multiple portions, thereby being suitable for SL settings. Thus, although prime researchers in FSL still investigate a simple CNN architecture,
several existing works in medical applications train ViTs from scratch and generate model improvement. \cite{park2021federated,park2022multi,almalik2023fesvibs,kim2022task}.

\textbf{Motivation.} Due to immense resource requirements when training ViTs from scratch in FSL, especially in a resource-constrained environment. Meanwhile, to the best of our knowledge, there is no previous research systematically analyzing the model performance under various heterogeneous data splits while considering local computing power and model privacy such as gradient inversion attacks \cite{wang2019beyond,huang2021evaluating} or black-box scenarios \cite{roumeliotis2023chatgpt, lin2023efficient} where gradient information or model parameters are inaccessible.

\textbf{Contribution.} In this paper, we aim to take the first investigation of systematically evaluating the performance of Pre-trained Image Transformer (PIT) in FSL scenarios. The main contributions are summarized as:
\begin{itemize}

\item We incorporate the pre-trained Image Transformer (PIT) into PSL scenarios, called FedV. To further enhance gradient information protection and have the capability compatible with black-box scenarios, we propose  FedVZ, which leverages zeroth-order (ZO) optimization into FedV.
\item We are the first to evaluate FSL performance with multiple PIT models in terms of model accuracy and convergence under various heterogeneous data distributions in many real-world datasets.
\item For the systematical comparison and evaluation, we conduct comprehensive experiments on CIFAR-10, CIFAR-100, and Tiny-ImageNet datasets with various PITs in terms of ViTs model family (ViT(S), ViT(T)) and distilled DeiTs model family (DDeiT(S), DDeiT(T), DDeiT(B)) as well as DeiTs model family (DeiT(S), DeiT(T), DeiT(B)) in non-IID settings with different data partitions, such as Dirichlet settings with various $\alpha$ and pathological settings with various limited classes in each client.
\end{itemize}
% 1. 第一次考虑在resource-constrained environment下对pre-trained Image transformer在FSL下性能做系统性测试评估【不同数据异质情况、convergence time、local resource overhead】
% 2. 同时考虑到local computing power【training overhead】、model privacy、model performance【pre-trained llms】；further enhance gradient information or have the capability compatible for black-box scenarios：incorporate ZO Optimization into FedV

\section{Related work}
\textbf{Federated Split Learning (FSL).} 
\noindent Federated split learning (FSL) \cite{thapa2021advancements} combines the strengths of two compelling distributed machine learning approaches: FL and SL paradigms. It trains the global model in parallel among distributed clients in FL and simultaneously splits the whole model as several sub-models between clients and the server in SL. Specifically, Splitfed \cite{thapa2022splitfed} is proposed to eliminate the inherent drawbacks of both FL and SL, which is the first training method in FSL. MiniBatch-SFL is presented in \cite{huang2023minibatch}, which incorporates MiniBatch SGD into FSL, where the clients train the client-side
model in a FL fashion while the server trains the server side and they also analyze the convergence on the non-IID data. \cite{mu2023communication} utilize an auxiliary network to locally update the client models while keeping only a single model on the server to alleviate the issue of high communication overhead. \cite{han2021accelerating} propose a new direction to FL/SL based on updating the client/server-side models in parallel via local-loss-based training specifically geared
to SL.
In addition, \cite{singh2019detailed} compare FL and SL settings under which each method outperforms the other in terms of communication efficiency. Meanwhile, \cite{gao2020end} provide empirical comparisons of FL and SL in real-world Internet of Things (IoT) settings in terms of learning performance and device implementation overhead.

% \textbf{Federated Learning with Image Transformers.}
% 1. FL with ViT is better than FL with CNN models;

% 2. works: FL with ViT;

\noindent
\textbf{Federated Split Vision Transformer.} 
To maintain data privacy in medical imaging scenarios while taking up less extensive computation resources and
bandwidth, recent works focus on the suitability of Transformers for collaborative learning. For instance,
FESTA \cite{park2021federated} is the first method for medical imaging classification tasks with a federated split vision transformer. Then, FeSViBS \cite{almalik2023fesvibs} builds upon the FeSTA framework and introduces a block sampling module, which leverages intermediate features extracted by the Vision Transformer (ViT) at the server. \cite{park2022multi} present a multi-task distributed learning framework p-FESTA using ViT with random patch permutation instead of using a CNN-based head as in FESTA. \cite{kim2022task} enable decomposition between the task-specific and common representations via an alternating training strategy between clients and server.

\noindent
\textbf{FL with Zeroth-order Optimization.}
Zeroth-order (ZO) optimization \cite{liu2020primer} is a subset of gradient-free optimization that emerges in many signal processing and machine learning (ML) applications with only function evaluations rather than requiring the gradient. In FL, \cite{fang2022communication} propose a derivative-free federated ZO optimization (FedZO) algorithm featured by performing multiple local updates based on stochastic gradient estimators in each communication round and enabling partial device participation. \cite{li2021communication} consider a local ZO algorithm based on biased stochastic ZO update in a decentralized FL setting. And these works \cite{zhang2022zeroth, tang2020distributed, zhang2021convergence} provide
a detailed analysis of their convergence behavior in distributed training.

The most related works to this paper are FESTA \cite{park2021federated}, FeSViBS \cite{almalik2023fesvibs}, FedZO \cite{fang2022communication}, Splitfed \cite{thapa2022splitfed}, and MiniBatch-SFL \cite{huang2023minibatch}.
However, these works still suffer from enormous resource overhead due to training LLMs from scratch (i.e., FESTA and FeSViBS) and inferior performance without leveraging the power of LLMs (i.e., Splitfed and MiniBatch-SFL), especially on heterogeneous data. Therefore, in this paper, we try to divide the PIT model pre-trained by public data into three parts. Specifically, we reside the PIT encoder on the server side, the head and tail networks are allocated in each client. Finally, these modules are updated simultaneously on the private data of each client. Furthermore, to consider the model privacy and the ability compatible with black-box scenarios, we utilize ZO optimization to approximate the gradient of the server model. Finally, we are the first to systematically evaluate the performance of our algorithms in terms of model accuracy and convergence under various heterogeneous data distributions in many real-world datasets.

% Overall, our work is unique ....
% \input{2_pre}
\section{Methodology}
In this section, we first define the problem setup for FSL with pre-trained image Transformers. After that, we elaborate our algorithms: FedV and  FedVZ.
\subsection{Problem Setup}
We consider a typical setting of FSL with $M$ clients, where each client~$i$ has the data distribution $\mathcal{D}_i$.  
Let $w\in\mathbb{R}^d$ represent the parameters of a pre-trained transformer model and $f_i(w;\xi_c)$ is the local objective function associated with the training data samples $\xi_c$. We decompose the whole transformer model as three parts in terms of the head $w_{\mathcal{H}}^i$ and tail $w_{\mathcal{T}}^i$ in local client $i \in \{1,2,...,M\}$ as well as transformer encoder $w_{\mathcal{E}}$ in the server side, which means $w=(w_\mathcal{H},w_\mathcal{E}, w_\mathcal{T} )$. 

After that, a common objective of FSL is the following finite-sum stochastic non-convex minimization problem:\\
\begin{equation}\label{dec}
    \min_{w} F(w) = \frac{1}{M}\sum_{i=1}^M \mathbb{E}_{\xi_c^{i,t} \sim \mathcal{D}_i } f_i(w; \xi_c).
\end{equation}

To be specific, the objective function of each client $i$ to optimize is
\begin{equation}\label{local_objective}
    \min_{w_\mathcal{H},w_\mathcal{T}} \mathbb{E}_{x_c^{i,t} \sim \mathcal{D}_i } l_c(y_c^{i,t}, \mathcal{T}_i(w_{\mathcal{T}}^i;\mathcal{E}(w_{\mathcal{E}}; \mathcal{H}_i(w_{\mathcal{H}}^i;\xi_c^{i,t})))).
\end{equation}
Where $\mathcal{H}_i$ and $\mathcal{T}_i$ represent the head network and tail network in the local client, while the transformer encoder is viewed as $\mathcal{E}$. The local training data is $\{(\xi_c^{i,t}, y_c^{i,t})\}$ in the $t$-th communication round.

% \begin{equation}
%     \min_{w_{\mathcal{H}},w_{\mathcal{T}}} \mathbb{E}_{\xi_c^{i,t} \sim \mathcal{D}_i } \mathcal{T}(w_{\mathcal{T}}^i;\mathcal{E}(w_{\mathcal{E}}; \mathcal{H}(w_{\mathcal{H}}^i;\xi_c^{i,t}))).
% \end{equation}
Meanwhile, for the transformer encoder update on the server side, the optimization problem is
\begin{equation}\label{server_objective}
    \min_{w_\mathcal{E}} \frac{1}{M}\sum_{i=1}^M \mathbb{E}_{x_c^{i,t} \sim \mathcal{D}_i } l_c(y_c^{i,t}, \mathcal{T}_i(w_{\mathcal{T}}^i;\mathcal{E}(w_{\mathcal{E}}; \mathcal{H}_i(w_{\mathcal{H}}^i;\xi_c^{i,t})))).
\end{equation}

% \begin{equation}
%     \min_{w_{\mathcal{E}}} \frac{1}{M}\sum_{i=1}^M \mathbb{E}_{\xi_c^{i,t} \sim \mathcal{D}_i } \mathcal{T}(w_{\mathcal{T}}^i;\mathcal{E}(w_{\mathcal{E}}; \mathcal{H}(w_{\mathcal{H}}^i;\xi_c^{i,t}))).
% \end{equation}
Therefore, for the whole model weights $w=(w_\mathcal{H},w_\mathcal{E}, w_\mathcal{T} )$, where $(w_\mathcal{E}, w_\mathcal{T} ) = (\frac{1}{M}\sum_{i=1}^M w_\mathcal{E}^i, \frac{1}{M}\sum_{i=1}^M w_\mathcal{T}^i)$. The objective function $f_i(w; \xi_c)$ in the problem \eqref{dec} can be represented as $l_c(y_c^{i,t}, \mathcal{T}_i(w_{\mathcal{T}}^i;\mathcal{E}(w_{\mathcal{E}}; \mathcal{H}_i(w_{\mathcal{H}}^i;\xi_c^{i,t}))))$. 
% \begin{equation}
%     \min_{w} F(w) = \frac{1}{M}\sum_{i=1}^M \mathbb{E}_{x_c^{i,t} \sim \mathcal{D}_i } f_i(w; \xi_c).
% \end{equation}
% where $f_i(w; \xi_c)=\mathcal{T}(w_{\mathcal{T}}^i;\mathcal{E}(w_{\mathcal{E}}; \mathcal{H}(w_{\mathcal{H}}^i;\xi_c^{i,t})))$

\textbf{The Challenge of FSL.} In FSL, due to the whole model being decomposed into several parts, the collaboration between these small models plays an essential role in achieving better performance after the FL training process.
\begin{algorithm*}
\small
\caption{FedV and  FedVZ}
\label{ FedVZ}
% \centering
\SetKwData{Left}{left}\SetKwData{This}{this}\SetKwData{Up}{up} \SetKwFunction{Union}{Union}\SetKwFunction{FindCompress}{FindCompress}
\SetKwInOut{Input}{Input}\SetKwInOut{Output}{Output}
\Input{Total number of clients $M$, sampling ratio of clients $q$, total number of communication rounds $T$, learning rate $\eta$, and total number of the local iterates are $K$. For gradient approximation, the perturbation scale is $\epsilon$ and randomly sampled $z \in \mathbb{R}^d $ with $z \sim \mathcal{N}(0, \mathbf{I}_d)$.} 
\Output{Generate the weights of local head and tail pairs $(w_{\mathcal{H}}^i, w_{\mathcal{T}}^i)$ for each client $i$.}
\textbf{Initialization:} Randomly initialize the local weights $(w_{\mathcal{H}}^i, w_{\mathcal{T}}^i)$ of the head and tail pairs for each client $i$. Meanwhile, the weights of the pre-trained body model $w_{\mathcal{E}}^0$ keep frozen and start to update when the FL training process begins. \\
% \BlankLine
\For{$t=0$ \KwTo $T-1$}{
    Sample a set of $m=qM $ clients at random without replacement, denoted by $\mathcal{W}^t $.\\
    \For{$k=1$ \KwTo $K$ }{
    \For{client $i=1$ \KwTo $m$ \emph{\textbf{in parallel}} }{
        $h_c^{k,t}(i) \gets  \textbf{ClientHead}(i, t) $. \Comment{Extract the intermediate feature from local data}
        
        $b_s^{k,t}(i) \gets \mathcal{E}(h_c^{k,t}(i))$. \Comment{Generate the smashed feature by using the server body model}
        
        $\mathcal{L}_c^{k,t}(i) \gets  \textbf{ClientTail}(i, b_s^{k,t}(i)) $. \Comment{Finish the forward pass and get the prediction}

        $(w^{i, k+1}_{\mathcal{H}}, w^{i, k+1}_{\mathcal{T}}) \gets \textbf{ClientUpdate}(i, \frac{\partial \mathcal{L}_c^{k,t}(i)}{\partial h_c^{k,t}(i)} )$.   \Comment{Start to backward propagation}

        \If{$k = K$}
        {
        \color{blue}{Option I: (FedV)} Generate and store the server gradient by averaging gradients: \color{black}{$g_s^{t}(i) =  \frac{\partial \mathcal{L}_c^{K,t}(i)}{\partial w_{\mathcal{E}}^t} $.}
        
        \color{blue}{Option II: ( FedVZ)} Generate and store the server gradient approximation by adopting ZO optimization:\\
        \color{black}{$g_s^{t}(i) \doteq  \frac{\mathcal{L}_c^{K,t}(i)(w_{\mathcal{E}}^{i, t} + \epsilon z) - \mathcal{L}_c^{K,t}(i)(w_{\mathcal{E}}^{i, t} - \epsilon z)}{2\epsilon}z  $. }
        }
        }}
        Update the weights of the server model for fine-tuning: $w_{\mathcal{E}}^{i, t+1} \gets w_{\mathcal{E}}^{i, t} - \frac{\eta}{m}\sum_{i=1}^m g_s^{t}(i) $.

        \If{\emph{t \% fedround = 0}}{Update the head/tail weights: $(w_{\mathcal{H}}^i, w_{\mathcal{T}}^i) \gets (\frac{1}{m}\sum_{i=1}^m w_{\mathcal{H}}^i, \frac{1}{m}\sum_{i=1}^mw_{\mathcal{T}}^i)$.}
}
\end{algorithm*}

\begin{algorithm}
% \small
\caption{Client-side Process}
\label{client_process}
\texttt{/* Runs on client $i$ in $k$-th local iteration at $t$-th round */}\\
\textbf{ClientHead:}\\
$\xi_c^{k,t}(i) \gets $ Sample minibatch data from client $i$.\\
 $h_c^{k,t}(i) \gets  \mathcal{H}_i(\xi_c^{k,t}(i)) $.

\textbf{Return} $h_c^{k,t}(i)$.

\textbf{ClientTail:}\\
$y_c^{k,t}(i) \gets $ The minibatch of the label from client $i$.\\
Compute loss value $\mathcal{L}_c^{k,t}(i) \gets  l_c(y_c^{k,t}(i), \mathcal{T}_i(b_s^{k,t}(i)))$.

\textbf{Return} $\mathcal{L}_c^{k,t}(i)$.

\textbf{ClientUpdate:}  \\
$w_{\mathcal{H}}^{i, k+1} \gets w_{\mathcal{H}}^{i,k} - \eta \frac{\partial \mathcal{L}_c^{k,t}(i)}{\partial w_{\mathcal{H}}^{i,k}}$. \Comment{Head back-propagation}

$w_{\mathcal{T}}^{i, k+1} \gets w_{\mathcal{T}}^{i,k} - \eta \frac{\partial \mathcal{L}_c^{k,t}(i)}{\partial w_{\mathcal{T}}^{i,k}}$. \Comment{Tail back-propagation}

\textbf{Return} $(w^{i, k+1}_{\mathcal{H}}, w^{i, k+1}_{\mathcal{T}}) $.
\end{algorithm}

\subsection{Algorithmic Details}
In this subsection, we present our two algorithms: FedV and  FedVZ. Our goal is to simultaneously optimize a server-side model (an encoder $\mathcal{E}$ of a pre-trained image transformer) and multiple client-side models in terms of head $\mathcal{H}_i$ and tail $\mathcal{T}_i$ networks to make local clients $i \in \{1,2,...,M\}$ capable of training the large transformer model under the constrained edge resources. Meanwhile, local models suitable for different heterogeneous data can benefit from the pre-trained image transformer, thereby accelerating the training process and reducing the training overhead. 
Below, we elaborate on the detailed procedure of our algorithms. The paradigm is introduced in Algorithm \ref{ FedVZ} and \ref{client_process}.

\textbf{Pre-trained Image Transformers (PITs) on Public Data.} Due to huge resource requirements when training ViTs from scratch in FSL, especially in a resource-constrained environment. Therefore, we utilize the public data with a similar feature to private data and then train the PITs on the private data of each client to reduce the training overhead in the FSL process.

\textbf{Forward Propagation.} During the forward propagation in both FedV and  FedVZ algorithms, at the beginning of each communication round, each client random initializes the local weights $(w_{\mathcal{H}}^i, w_{\mathcal{T}}^i)$ of the head and tail pairs and the encoder weights $w_{\mathcal{E}}$ at the server side retain the weights of the pre-trained image transformer model. Then several clients are sampled to participate in this round and perform their local iteration updates. For each sampled client $i$, it calculates the intermediate feature $h_c^{k,t}(i)$ from its local minibatch data by using their local head model $\mathcal{H}_i$, which means each client executes the forward propagation. After that, all participating clients transmit their intermediate features to the server side. 

When the pre-trained transformer encoder obtains these local features on the server side, the encoder performs forward propagation to generate the server-side features $b^{k,t}_s(i)$ for corresponding client $i$ in parallel. Finally, the server-side feature $b^{k,t}_s(i)$ is sent out to local client $i$ simultaneously, and then the local tail network $\mathcal{T}_i$ calculates the final result $\mathcal{T}_i(b_s^{k,t}(i)))$ and loss value $\mathcal{L}_c^{k,t}(i)$ according to the local data in parallel.

\textbf{Backward Propagation.} Different from previous works, we adopt a pre-trained transformer encoder to accelerate the training process and only utilize the loss value $\{\mathcal{L}_c^{K,t}(1), \mathcal{L}_c^{K,t}(2),..., \mathcal{L}_c^{K,t}(m)\}$ of all participated clients $i \in \{1,2,...,m\}$ at the final local iteration step $K$ in each communication round $t$ while having similar performance. To massively reduce the communication overhead and local resources cost in terms of computation and memory overheads, it is a significant improvement relative to almost all existing FSL methods, such as classic methods: SplitFed, minibatch-FSL, and FeSViBS.

To be specific, for FedV, the target is to minimize the objective function \eqref{local_objective}. So for client $i$, we can generate the gradients of the local head and tail network, which is  $\frac{\partial \mathcal{L}_c^{k,t}(i)}{\partial w_{\mathcal{H}}^{i,k}}$ and $\frac{\partial \mathcal{L}_c^{k,t}(i)}{\partial w_{\mathcal{T}}^{i,k}}$, respectively. After that, the gradients of these local gradients are transmitted to the server. Subsequently, the transformer encoder computes the averaged gradient:
\begin{equation}\small
    \frac{1}{m}\sum_{i=1}^m g_s^{t}(i) =  \frac{1}{m}\sum_{i=1}^m\frac{\partial \mathcal{L}_c^{K,t}(i)}{\partial w_{\mathcal{E}}^t} .
\end{equation}
Note the loss value $\mathcal{L}_c^{K,t}(i)$ is related to the local head weights $w_{\mathcal{H}}^{i,K}$ and tail model weights $w_{\mathcal{T}}^{i,K}$ of client $i$, where $K$ is the final local iteration step. Then, it performs one step of gradient update.

For  FedVZ, the main focus is on the privacy of the model weights and gradient information in practice, such as black-box scenarios, we adopt the zero-order perturbation following the Simultaneous Perturbation Stochastic Approximation (SPSA) \cite{spall1992multivariate,spall1997one} to approximate the high-dimensional gradient efficiently.
For gradient approximation, let the perturbation scale is $\epsilon \in (0, 1)$ and randomly sampled is $z \in \mathbb{R}^d $ with $z \sim \mathcal{N}(0, \mathbf{I}_d)$. Thus, we obtain the server gradient by:
\begin{equation}
    g_s^{t}(i) \doteq  \frac{\mathcal{L}_c^{K,t}(w_{\mathcal{E}}^{i, t} + \epsilon z) - \mathcal{L}_c^{K,t}(w_{\mathcal{E}}^{i, t} - \epsilon z)}{2\epsilon}z  .
\end{equation}

Note that the primary backward propagation procedure is identical to that in FedV. The sole difference is how to estimate the transformer encoder gradient on the server side due to some important concerns, such as model privacy.

\textbf{Client-side Model Aggregation.} For both algorithms, after several communication rounds, the client-side head and tail parts need to be sent out to the server for aggregating and updating the model weights as the same as the general FL methods. In practice, we find that this parameter is a significant factor in deciding the training process, thereby affecting the final model performance.

\section{Experiments}
\subsection{Experimental Setup}

\noindent
\textbf{Dataset and Data Partition.} We evaluate our approaches on CIFAR-10, CIFAR-100  \cite{krizhevsky2009learning}, and Tiny-ImageNet  \cite{le2015tiny} datasets with Dirichlet and Pathological data partition.  All detailed experiments
on Tiny-ImageNet are placed in \textbf{Appendix} \ref{more_exper} due to the limited space.
% All detailed experiments on Tiny-ImageNet are placed in \textbf{Appendix} \ref{exper:tiny} due to the limited space. 
For evaluating performance in various non-IID scenarios, we partition the training and testing data according to the Dirichlet distribution Dir($\alpha$) and Pathological data partition Path-c. Note that the number of sampling classes is represented as “c”. For instance, we sample 2 and 5 classes from a total of 10 classes on CIFAR-10, and 5 and 10 classes from a total of 100 classes on CIFAR-100 respectively for each client. 
For the Dir($\alpha$) partition, we set $\alpha = \{ 0.1, 0.3\}$ for all datasets, where the smaller the $\alpha$ is, the more heterogeneous the setting is. For the Path-c partition, we set $c = \{2,5,10\}$ in different datasets, where the fewer classes $c$ each client has, the more heterogeneous the setting is.

\noindent
\textbf{Models and Baselines.} 
We use various PITs in terms of ViTs model family (ViT(S), ViT(T)) and distilled DeiTs model family (DDeiT(S), DDeiT(T), DDeiT(B)) as well as DeiTs model family (DeiT(S), DeiT(T), DeiT(B)) for all datasets in various non-IID settings with different data partitions to verify the effectiveness of our algorithms. For the systematical comparison, we compare the proposed approaches with: 1) SOTA FL baselines in terms of FedAvg \cite{mcmahan2017communication}, Fed-RoD \cite{chen2021bridging}, and Ditto \cite{li2021ditto}; 2) SOTA FSL baseline: FESTA \cite{park2021federated}.
All baselines using the ResNet model family \cite{he2016deep} replace the batch normalization with the group normalization to avoid unstable performance.

Furthermore, to evaluate the performance of our algorithms fairly compared with baselines using CNN models, we find that DeiT(S) is similar to ResNet-50 in terms of parameter size (MB) and FLOPs (B) as well as ImageNet result measured by Table 2 in \cite{han2022survey}. The compared table is shown below. Thus, we use DeiT(S) for all algorithms using transformers and ResNet-50 for all baselines using CNN models in the same setting, respectively.

\begin{table}
\centering
\caption{Comparison between ResNet-50 and DeiT(S). Note that Top-1 (\%) accuracy is the result on ImageNet and the throughput is measured on NVIDIA V100 GPU and Pytorch, with $224 \times 224$ input size.}
\label{ta:model_compare}
\begin{tabular}{ccccc} 
\toprule
Model     & Paras (MB) & FLOPs (B) & Throughput (image/s) & Top-1 (\%)  \\ 
\midrule
ResNet-50 & 25.6      & 4.1       & 1226                 & 79.1        \\ 
\midrule
DeiT(S)   & 22        & 4.6       & 940                  & 79.8        \\
\bottomrule
\end{tabular}
\end{table}

\noindent
\textbf{Implementation Details.}
We perform 500 rounds with 100 clients on CIFAR-10, CIFAR-100, and Tiny-ImageNet with different client sampling ratios. The batch size is 128. We set SGD \cite{robbins1951a} as the base optimizer of baselines with a learning rate of 0.1 and local momentum of $0.9$. We report the mean performance with 3 different seeds. More details can be found in \textbf{Appendix} \ref{exp:hyperparameters}.

\begin{table*}
\centering
\scriptsize
\caption{ Test accuracy (\%) on CIFAR-10 \& 100 in both Dirichlet and Pathological distribution settings.}
\label{ta:all_baselines}
\vspace{0.1cm}
\scalebox{1}{\begin{tabular}{lcccc|cccc} 
\toprule
\multirow{3}{*}{Algorithm} & \multicolumn{4}{c|}{CIFAR-10}                                                                                          & \multicolumn{4}{c}{CIFAR-100}                                                                                            \\ 
\cmidrule{2-9}
                           & \multicolumn{2}{c}{Dirichlet}                            & \multicolumn{2}{c|}{Pathological}                           & \multicolumn{2}{c}{Dirichlet}       & \multicolumn{2}{c}{Pathological}                                                   \\ 
\cmidrule{2-9}
                           & $\alpha$ = 0.1               & $\alpha$ = 0.3            & c = 2                        & c = 5                        & $\alpha$ = 0.1   & $\alpha$ = 0.3   & c = 5                                                  & c = 10                    \\ 
\midrule

FedAvg                     & $86.17_{\pm.28}$               & $82.02_{\pm.20}$            & $86.99_{\pm.11}$               & $83.18_{\pm.27}$               & $59.35_{\pm.03}$   & $57.12_{\pm.06} $  & $ 71.29_{\pm.43} $                                       & $68.10_{\pm.48} $           \\

Fed-RoD                    & $91.15_{\pm.12}$               & $87.68_{\pm.08}$            & $92.10_{\pm.04} $              & $89.81_{\pm.45}$                        & $ 67.79_{\pm.05} $ & $60.54_{\pm.69} $  & $ 82.50_{\pm.45} $                                       & $ 75.59_{\pm.15} $          \\
Ditto                      & $82.22_{\pm.10}$               & $75.51_{\pm.04}$            & $86.96_{\pm.40} $              & $ 77.59_{\pm.32} $             & $50.85_{\pm.54 }$  & $50.65_{\pm.50} $  & $71.48_{\pm.45} $                                        & $ 62.77_{\pm.30} $          \\ 
\midrule
FESTA                  & $75.16_{\pm.53}$               & $ 65.73_{\pm.30} $          & $86.53_{\pm.55 }$              & $75.61_{\pm.17} $ & $ 40.98_{\pm.10}$  & $ 78.40_{\pm.31} $ & $69.58_{\pm.15} $                                        & $ 71.08_{\pm.52} $          \\
\midrule
FedV                   & $96.28_{\pm.13}$               & $\textbf{96.87}_{\pm.20} $           & $94.77_{\pm.40} $              & $\textbf{97.54}_{\pm.25} $              &   $\textbf{85.55}_{\pm.46} $                & $\textbf{81.49}_{\pm.25}$                 &        $93.71_{\pm.75} $                     &      $93.10_{\pm.17} $        \\
 FedVZ                   & $\textbf{96.93}_{\pm.10}$      & $ 95.88_{\pm.12} $ & $ \textbf{96.63}_{\pm.30} $    & $96.85_{\pm.21} $   & $ 83.05_{\pm.41} $ & $ 80.51_{\pm.13} $    & $\textbf{96.77}_{\pm.30} $    &  $\textbf{94.02}_{\pm.25}$   \\
\bottomrule
\end{tabular}}
\vspace{-0.3cm}
\end{table*}
\begin{figure}
	\centering
        \begin{subfigure}{1\linewidth}
        \centering
            \includegraphics[width=1\textwidth]{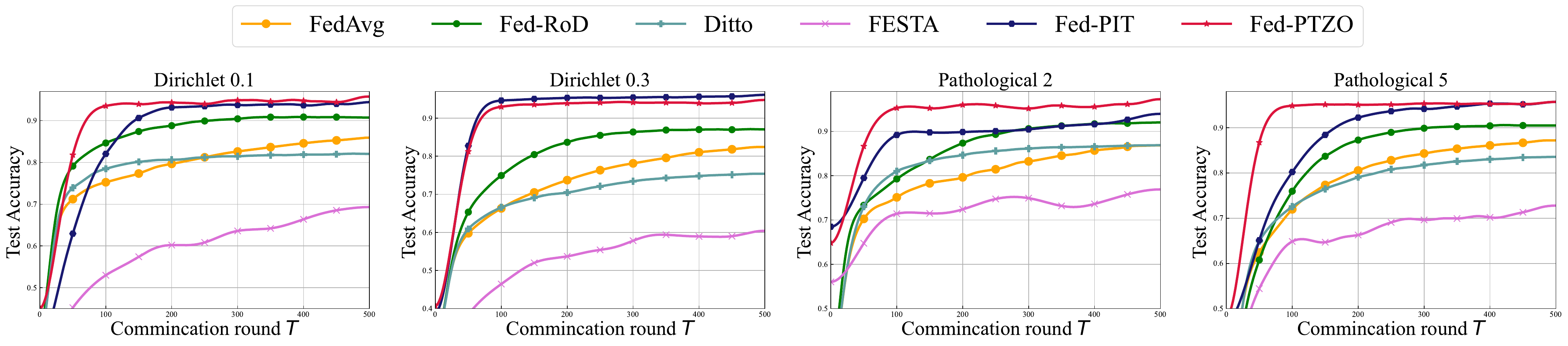}
            \caption{CIFAR-10}
            \label{fig:cifar10}
        \end{subfigure}
	  \begin{subfigure}{1\linewidth}
        \centering
            \includegraphics[width=1\textwidth]{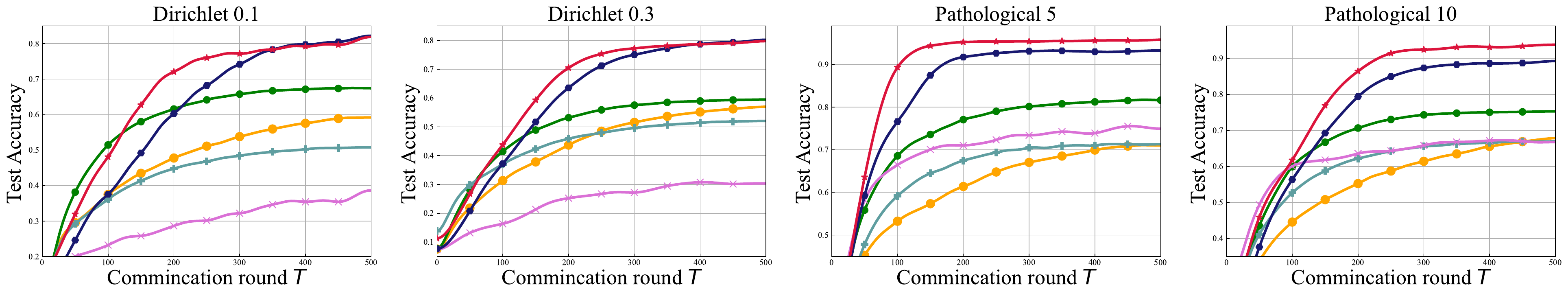}
            \caption{CIFAR-100}
            \label{fig:cifar100}
        \end{subfigure}
        \caption{\small The averaged testing accuracy on \textit{CIFAR-10} and \textit{CIFAR-100} dataset for all compared methods. }
 \label{fig:all}
\end{figure}

\subsection{Performance Evaluation}\label{exper:bas}

\textbf{Comparison with the baselines.}
As shown in Table \ref{ta:all_baselines} and Figure \ref{fig:all}, the proposed FedV and  FedVZ outperform other baseline methods with the best stability and better performance in both different datasets and different data heterogeneity scenarios. Specifically, on the CIFAR-10 dataset, FedV and  FedVZ achieve 96.87\% and 95.88\% on the Directlet-0.3 setting, 9.19\% and 8.20\% ahead of the best-compared method Fed-RoD. On the CIFAR-100 dataset,  FedVZ achieves at least 14.27\% and 19.43\% improvement from the other baselines on the Pathological-5 and Pathological-10 settings. We attribute that FedV and  FedVZ have very superior performance and are considerably suitable for the FSL setting based on the performance advantage of PITs.
It is also observed that  FedVZ has better performance while protecting the model privacy on the server side, demonstrating that ZO optimization is capable of applying in distributed learning, such as FSL.

\noindent
\textbf{Evaluation on various non-IID scenarios.}
We discuss two types of data heterogeneity: Dirichlet distribution and Pathological distribution under various pre-trained transformer models on the CIFAR-10 dataset in Table \ref{ta:non-iid}, and prove the effectiveness and robustness of the proposed methods. In Dirichlet distribution, since the local training can't cater for all classes inside clients, the accuracy decreases with the level of heterogeneity decreasing. It is clearly seen that our algorithms: FedV and  FedVZ
can achieve stronger stability for several heterogeneous settings under various pre-trained transformer models. Pathological distribution defines limited classes for each client which is a higher level of heterogeneity. FedV and  FedVZ still have better performance on CIFAR-10 with only 2 categories per client.

\begin{table}
\centering
\caption{ Test accuracy (\%) on various Dirichlet and Pathological distribution settings with different PIT models.}
\label{ta:non-iid}
\scriptsize
\begin{tabular}{ccccc|ccc|cc} 
\toprule
\multirow{3}{*}{Algorithm} & \multirow{3}{*}{Non-IID} & \multicolumn{8}{c}{Model}                                                                              \\ 
\cmidrule{3-10}
                           &                          & \multicolumn{3}{c|}{Distilled-DeiT}  & \multicolumn{3}{c|}{DeiT}            & \multicolumn{2}{c}{ViT}  \\ 
\cmidrule{3-10}
                           &                          & Tiny       & Small      & Base       & Tiny       & Small      & Base       & Tiny       & Small       \\ 
\midrule
\multirow{4}{*}{FedV}    & Dir-0.1                  & $81.45_{\pm.46}$ &$ 97.17_{\pm.06}$ &$ 97.89_{\pm.19}$ &$ 81.24_{\pm.20}$ &$ 96.29_{\pm.14}$ &$ 96.29_{\pm.14}$ &$ 80.36_{\pm.41}$ &      $ 94.64_{\pm.45}$       \\
                           & Dir-0.3                  &$ 95.28_{\pm.14}$ &$ 96.72_{\pm.18}$ &$ 97.40_{\pm.20}$ &$ 96.02_{\pm.16}$ &$ 96.97_{\pm.17}$ &$ 96.97_{\pm.17}$ &$ 74.82_{\pm.21}$ &$ 97.80_{\pm.30}$  \\
                           & Pat-2                    &$ 90.39_{\pm.72}$ &$ 98.20_{\pm.40}$ &$ 98.86_{\pm.20}$ &$ 91.19_{\pm.92}$ &$ 91.60_{\pm.79}$ &$ 91.60_{\pm.79}$ &$ 90.80_{\pm.38}$ &$ 92.38_{\pm.34}$  \\
                           & Pat-5                    &$ 95.87_{\pm.38}$ &$ 97.50_{\pm.25}$ &$ 98.15_{\pm.13}$ &       $ 96.91_{\pm.51}$     &$ 97.54_{\pm.25}$ &$ 97.54_{\pm.25}$ &$ 81.69_{\pm.77}$ &$ 97.73_{\pm.35}$  \\ 
\midrule
\multirow{4}{*}{ FedVZ}  & Dir-0.1                  &$ 95.72_{\pm.10}$ &$ 96.99_{\pm.09}$ &$ 98.30_{\pm.08}$ &$ 79.32_{\pm.24}$ &$ 97.05_{\pm.21}$ &$ 97.05_{\pm.21}$ &$ 78.36_{\pm.30}$ &$ 78.98_{\pm.18}$  \\
                           &Dir-0.3                  &$ 93.64_{\pm.42}$ &$ 95.96_{\pm.24}$ &$ 97.52_{\pm.40}$ &$ 71.88_{\pm.41}$ &$ 95.88_{\pm.17}$ &$ 95.88_{\pm.17}$ &$ 70.34_{\pm.41}$ &$ 71.66_{\pm.47 }$ \\
                           & Pat-2                    &$ 90.10_{\pm.60}$ &$ 98.42_{\pm.35}$ &$ 99.23_{\pm.22}$ &$ 89.72_{\pm.49}$ &$ 90.45_{\pm.15}$ &$ 90.45_{\pm.15}$ &$ 89.38_{\pm.28}$ &$ 90.28_{\pm.15}$  \\
                           & Pat-5                    &$ 95.68_{\pm.71}$ &$ 97.17_{\pm.14}$ &$ 98.37_{\pm.34}$ &$ 76.62_{\pm.12}$ &$ 96.85_{\pm.26}$ &$ 96.85_{\pm.26}$ &$ 75.51_{\pm.15}$ &$ 76.77_{\pm.28}$  \\
\bottomrule
\end{tabular}
\end{table}
\noindent
% \textbf{Convergence speed.}

\subsection{Ablation Study}
In this subsection, we verify the influence of the main hyper-parameters in Fed-PIT. All the ablation studies are conducted on the CIFAR-10 dataset.

\begin{figure}[htbp]
	\centering
	\begin{subfigure}[b]{.32\textwidth}
		\centering
		\includegraphics[width = \textwidth]{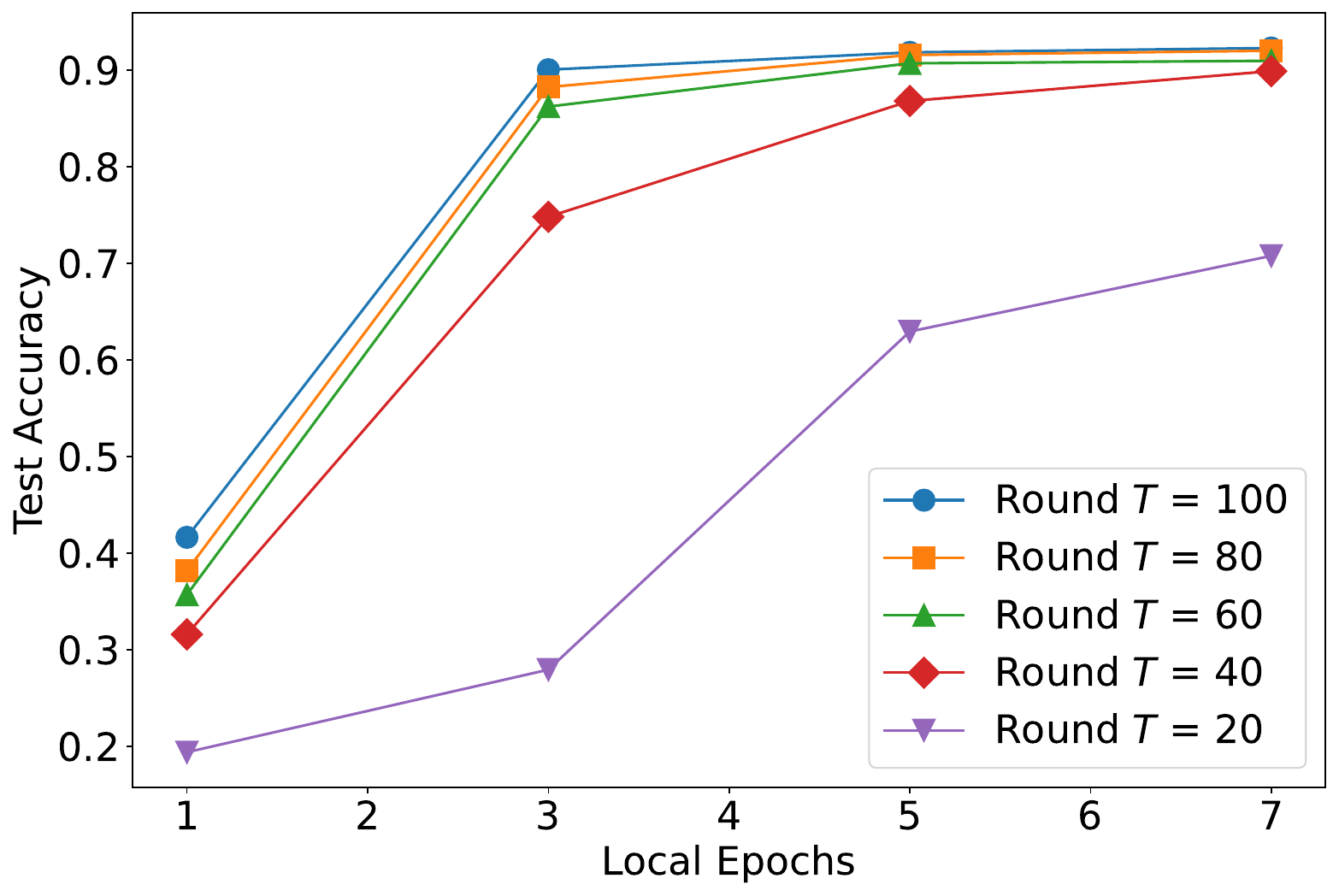}
		\caption{Effect of the local epochs.}
            \label{local_epoch}
	\end{subfigure}
	\begin{subfigure}[b]{.32\textwidth}
		\centering
		\includegraphics[width = \textwidth]{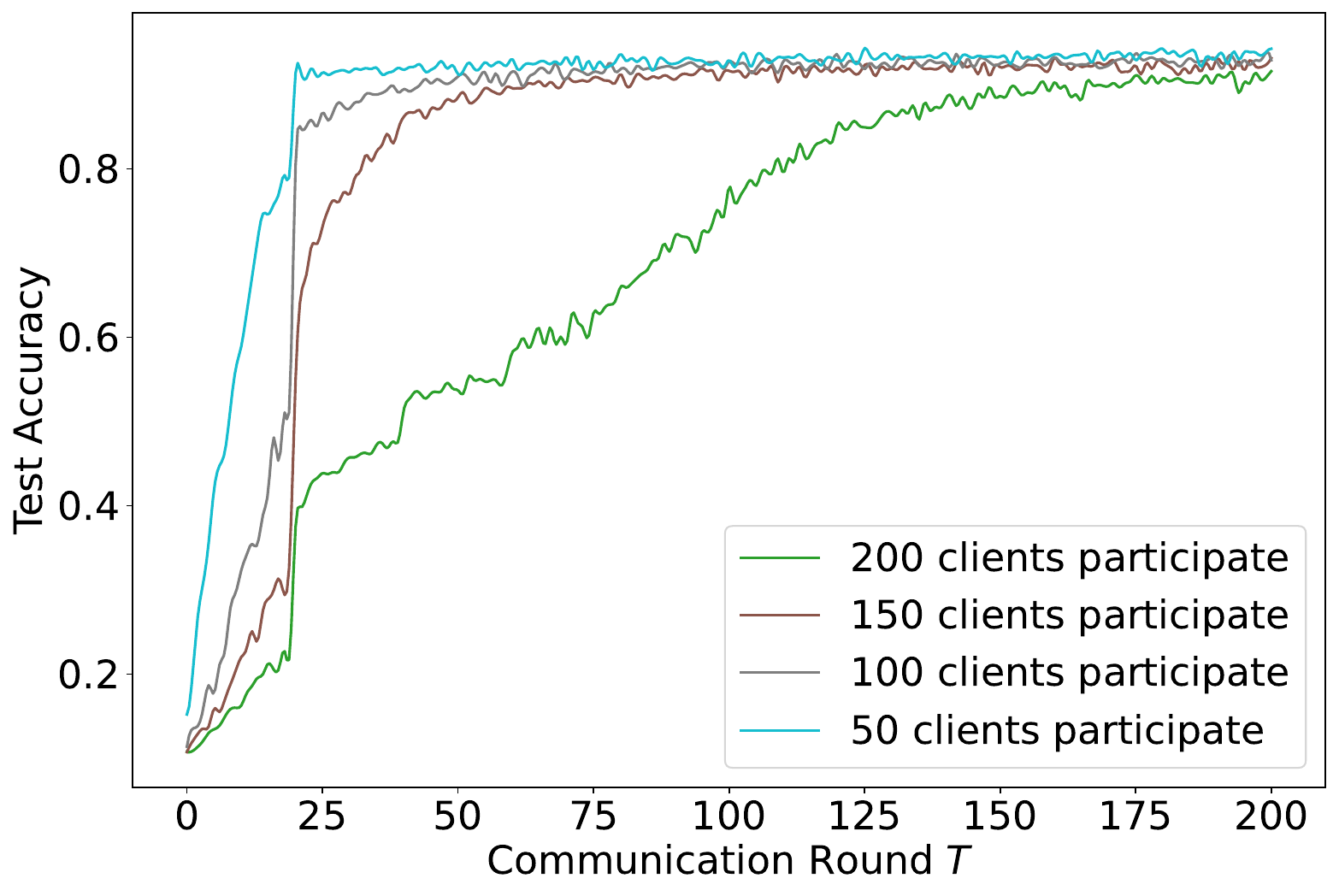}
		\caption{Effect of the clients' size.}
            \label{clients_num}
	\end{subfigure}
        \begin{subfigure}[b]{.32\textwidth}
		\centering
		\includegraphics[width = \textwidth]{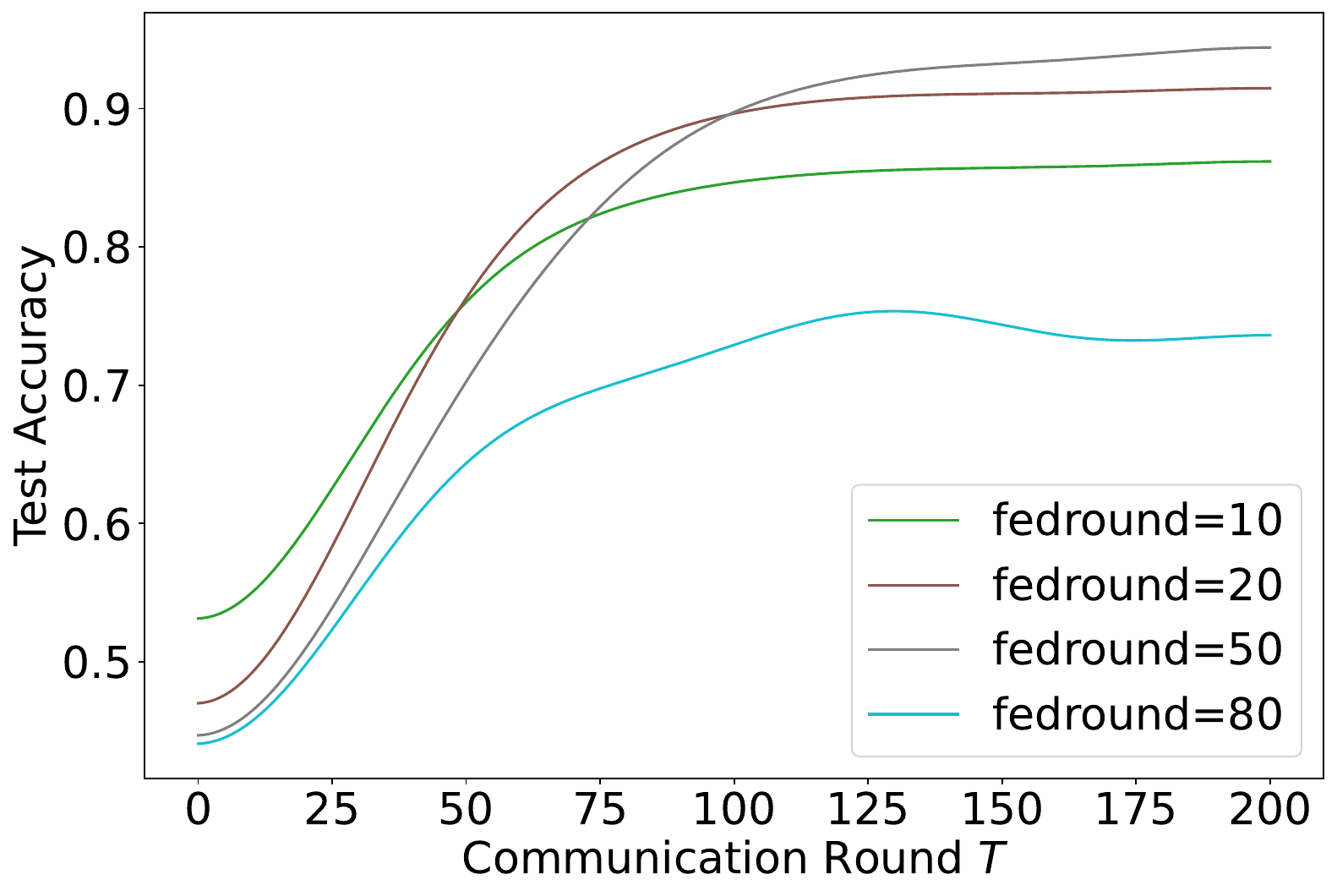}
		\caption{Effect of fedround.}
            \label{fedround}
	\end{subfigure}
 \caption{ Effect of the main hyper-parameters: local iteration steps $K$, the number of participated clients $M$, and the local model aggregation interval: fedround.}
\end{figure}

\noindent
\textbf{Local iteration steps $K$.}
In Figure \ref{local_epoch}, we illustrate the effect of local epochs in different communication rounds $T$ ($T = \{20, 40,60, 80, 100\}$) on the CIFAR-10. To investigate
the acceleration on $T$ by adopting a larger local iteration steps $K$, we fix the total batchsize and change local training epochs. With fixed communication round $T$, fewer local epochs of our method have inferior model performance. With the increase of local epochs, the accuracy is more superior. It is clear that under various communication rounds $T$, the testing accuracy achieves the best value when the local epoch is set to 5. Thus, in our paper, we set the local epoch to 5 for all algorithms.

\noindent
\textbf{Number of participated clients $M$.} We compare the model performance between different numbers of client participation on the CIFAR-10 with the data distribution of Dir-0.3 in Figure \ref{clients_num}. Compared with the larger participated clients \{150, 200\}, the smaller participated clients \{50, 100\} achieve better training convergence and testing accuracy due to the number of local training data increasing.

\noindent
\textbf{Local model aggregation interval: fedround.}
In Figure \ref{fedround}, we illustrate the effect of local model aggregation interval: fedround. It is obvious that the convergence speed is fastest when fedround is set to 20 and the test accuracy achieves the best performance when fedround is set to 50. To consider the convergence acceleration during the model aggregation, we set fedround to 20 in this paper.
\noindent
% \textbf{The main factors of ZO optimization.}
\section{Conclusion}
In this paper, we focus on the model privacy and the resource overhead issues as 
well as performance improvement when training LLMs in a resource-constrained environment, e.g., IoT. We are the first to investigate this area and systematically confirm the performance of training PIT models with the way of both distributed training and model splitting. For the systematical comparison and evaluation, we conduct comprehensive experiments on CIFAR-10, CIFAR-100, and Tiny-ImageNet datasets with various PITs in terms of  ViTs model family (ViT(S), ViT(T)) and distilled DeiTs model family (DDeiT(S), DDeiT(T), DDeiT(B)) as well as DeiTs model family (DeiT(S), DeiT(T), DeiT(B))in non-IID settings with different data partitions. We hope this work can provide some insights and give a feasible direction to further study harnessing the computational power of LLMs in the FSL community.

% \clearpage
\bibliographystyle{splncs04}
\bibliography{ref}

\clearpage
\newpage
\onecolumn 

\vspace{-1cm}
\begin{center}
 \rule{4.5in}{0.7pt}\\ % 4.0
 \vspace{0.25cm}
 {\Large\bf Supplementary Material for ``Heterogeneous Federated Learning with Splited Language Model''}
 \rule{4.5in}{0.7pt}
\end{center}
\appendix

\section{More Details in the Implementation.}\label{exp:hyperparameters}
\subsection{Datasets and Data Partition}

CIFAR-10/100 and Tiny-ImageNet are three basic datasets in machine learning. As shown in Table \ref{ta:all_data}, they are all colorful images with different classes and different resolutions. We use two non-IID partition methods to split the training data in our implementation. One is based on Dirichlet distribution on the label ratios to ensure data heterogeneity among clients. The Dirichlet distribution defines the local dataset to obey a Dirichlet distribution, where a smaller $\alpha$ means higher heterogeneity. Another assigns each client a limited number of categories, called Pathological distribution. Pathological distribution defines the local dataset to obey a uniform distribution of active categories $c$, where fewer categories mean higher heterogeneity. The distribution of the test datasets is as the same as in training datasets. We run 500 communication rounds for CIFAR-10, CIFAR-100, and 300 rounds for Tiny-ImageNet.
% \subsection{Communication Topology}

\begin{table}[ht]
\centering
\small
\caption{The details on three datasets.}
\label{ta:all_data}
\begin{tabular}{ccccc} 
\toprule
Dataset       & Training Data & Test Data & Class & Size     \\ 
\midrule
CIFAR-10      & 50,000        & 10,000    & 10    & 3×32×32  \\
CIFAR-100     & 50,000        & 10,000    & 100   & 3×32×32  \\
Tiny-ImageNet & 100,000       & 10,000    & 200   & 3×64×64        \\
\bottomrule
\end{tabular}
\end{table}

\subsection{More details about the  FedV and  FedVZ}\label{ap:hyperparameters}
For  FedV and  FedVZ, we train the model for 5 epochs per round as the same as other baselines. Therefore, we conduct ablation experiments to select the most appropriate parameters for each data distribution. We set fedround to 50. Model parameters are optimized by Adam with a learning rate decay of 0.0002. The weight perturbation ratio in  FedVZ is set to zero in the experiments. Furthermore, in  FedVZ, the main hyper-parameters of ZO optimization are set to $\epsilon = 0.0001$, the learning rate of the server-side model $lr_{server} = 1e-6$ with the decaying rate 0.5 and the learning rate of the client model $lr_{client} = 2e-4$.

\subsection{More Details about Baselines}\label{exp:2}

\textbf{FedAvg} \cite{mcmahan2017communication} is the most commonly discussed method in FL. It selects some clients to perform local training on each dataset and then aggregates the trained local models to update the global model. Actually, the local model in FedAvg is also the comparable personalized model for each client.

\noindent
\textbf{Fed-RoD} \cite{chen2021bridging} explicitly decouples a model’s dual duties with two prediction tasks: generic optimization and personalized optimization and utilizes a hyper network to connect the generic model and the personalized model. Each client first updates the generic model with balanced risk minimization then updates the personalized model with empirical risk minimization.

\noindent
\textbf{Ditto} \cite{li2021ditto} achieves personalization via a trade-off between the global model and local objectives. It totally trains two models on the local datasets, one for the global model (similarly aggregated as in FedAvg) with local empirical risk, and one for the personal model (kept locally) with both empirical and proximal terms towards the global model. We perform a grid search (0.1, 0.4, 0.75, 1) of the regularization parameters $\lambda$, then report the best 0.75 in our experiments. 

\noindent
\textbf{FESTA} \cite{park2021federated} is the first method for medical imaging classification tasks with a federated split vision transformer. We train the model for 5 epochs per round. Therefore, we conduct ablation experiments to select the most appropriate parameters for each data distribution. We set fedround to 50. Model parameters are optimized by Adam with a learning rate decay of 0.0002. The weight perturbation ratio is set to zero in the experiments.

\section{More Details in the Experiment.}\label{more_exper}

\begin{figure}
	\centering
    
            \includegraphics[width=1\textwidth]{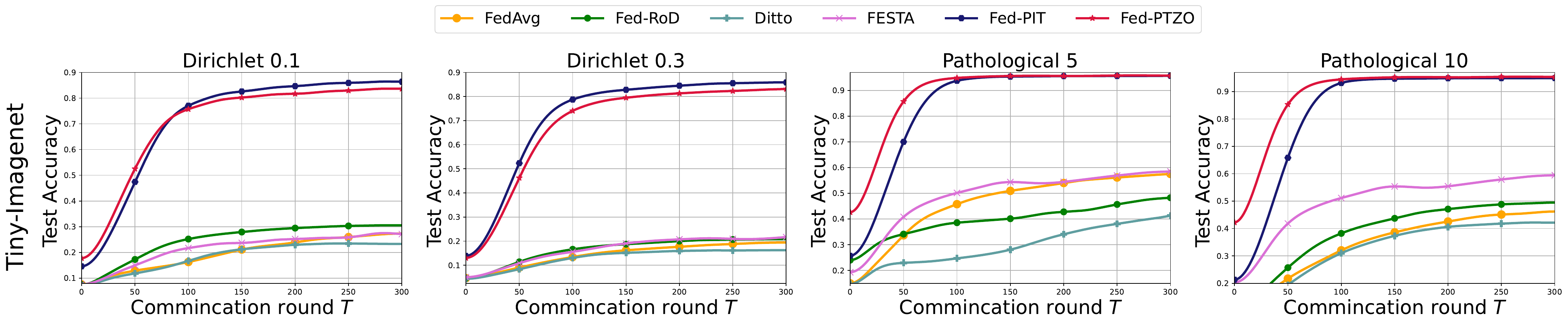}
            \caption{Tiny-ImageNet}
        \caption{\small The averaged testing accuracy on \textit{Tiny-ImageNet} dataset for all compared methods. }
 \label{fig:tiny}
\end{figure}

\begin{table}
\centering
\caption{ Test accuracy (\%) on Tiny-ImageNet in both Dirichlet and Pathological distribution settings.}
\label{ta:tiny_all_baselines}
\begin{tabular}{ccccc} 
\toprule
\multirow{3}{*}{Algorithm} & \multicolumn{4}{c}{Tiny-ImageNet}                                 \\ 
\cmidrule{2-5}
                           & \multicolumn{2}{c}{Dirichlet} & \multicolumn{2}{c}{Pathological}  \\ 
\cmidrule{2-5}
                           & $\alpha$ = 0.1         & $\alpha$ = 0.3           & c = 5        & c = 10                 \\ 
\midrule
FedAvg                     & 14.13 $\pm$ 0.13   & 7.42 $\pm$ 0.21      & 60.97 $\pm$ 0.33 & 46.56 $\pm$ 0.39           \\
Ditto                      & 23.71 $\pm$ 0.66   & 16.47 $\pm$ 0.14     & 49.84 $\pm$ 0.75 & 42.65 $\pm$ 0.15           \\
FedRoD                     & 31.03 $\pm$ 0.55   & 21.25 $\pm$ 0.45     & 56.42 $\pm$ 0.47 & 50.01 $\pm$ 0.40           \\
FESTA                      & 36.75 $\pm$ 0.73   & 27.10 $\pm$ 0.18     & 66.79 $\pm$ 0.50 & 65.45 $\pm$ 0.25           \\ 
\midrule
 FedV                  & \textbf{89.35} $\pm$ 0.30   & \textbf{87.77 }$\pm$ 0.08     & 96.77 $\pm$ 0.10 & 95.70 $\pm$ 0.50           \\
 FedVZ                     & 87.10 $\pm$ 0.11 & 85.40 $\pm$ 0.50   & \textbf{96.83} $\pm$ 0.16 & \textbf{96.41} $\pm$ 0.33           \\
\bottomrule
\end{tabular}
\end{table}

\noindent
\textbf{Comparison with the baselines.}
As shown in Table \ref{ta:tiny_all_baselines} and Figure \ref{fig:tiny}, the proposed FedV and  FedVZ outperform other baseline methods with the best stability and better performance in both different datasets and different data heterogeneity scenarios. Specifically, FedV and  FedVZ achieve 87.77\% and 85.40\% on the Directlet-0.3 setting, which are very far ahead of the best-compared method Fed-RoD. We attribute that FedV and  FedVZ have very superior performance and are considerably suitable for the FSL setting based on the performance advantage of PITs.

\noindent
\textbf{Evaluation on various non-IID scenarios.}
We discuss two types of data heterogeneity: Dirichlet distribution and Pathological distribution under various pre-trained transformer models in Table \ref{ta:tiny_all_baselines}, and prove the effectiveness and robustness of the proposed methods. In Dirichlet distribution, since the local training can't cater for all classes inside clients, the accuracy decreases with the level of heterogeneity decreasing. It is clearly seen that our algorithms: FedV and  FedVZ
can achieve stronger stability for several heterogeneous settings under various pre-trained transformer models.

\end{document}